\documentclass[11pt]{article}
\usepackage[utf8]{inputenc}
\usepackage[T1]{fontenc}
\usepackage[arabic, english]{babel} 
\usepackage[preprint]{acl}
\usepackage{comment}
\usepackage{times}
\usepackage{latexsym}

\usepackage[T1]{fontenc}

\usepackage[utf8]{inputenc}

\usepackage{microtype}

\usepackage{inconsolata}

\usepackage{graphicx}

\title{QU-NLP at QIAS 2025 Shared Task: A Two-Phase LLM Fine-Tuning and Retrieval-Augmented Generation Approach for Islamic Inheritance Reasoning}

\author{Mohammad AL-Smadi \\
  Qatar University \\
  Doha, Qatar \\
  \texttt{malsmadi@qu.edu.qa} \\}

\begin{document}
\maketitle

\begin{abstract}
This paper presents our approach and results for SubTask 1: Islamic Inheritance Reasoning at QIAS 2025, a shared task focused on evaluating Large Language Models (LLMs) in understanding and reasoning within Islamic inheritance knowledge. We fine-tuned the Fanar-1-9B causal language model using Low-Rank Adaptation (LoRA) and integrated it into a Retrieval-Augmented Generation (RAG) pipeline. Our system addresses the complexities of Islamic inheritance law, including comprehending inheritance scenarios, identifying eligible heirs, applying fixed-share rules, and performing precise calculations. Our system achieved an accuracy of 0.858 in the final test, outperforming other competitive models such as, GPT 4.5,  LLaMA, Fanar, Mistral and ALLaM evaluated with zero-shot prompting. Our results demonstrate that QU-NLP achieves near state-of-the-art accuracy (85.8\%), excelling especially on advanced reasoning (97.6\%) where it outperforms Gemini 2.5 and OpenAI's o3. This highlights that domain-specific fine-tuning combined with retrieval grounding enables mid-scale Arabic LLMs to surpass frontier models in Islamic inheritance reasoning.
\end{abstract}

\section{Introduction}


The rapid advancements in Large Language Models (LLMs) have opened new avenues for their application across diverse domains, including specialized knowledge systems. This paper details our participation in the QIAS~2025 Shared Task, specifically focusing on Subtask~1: Islamic Inheritance Reasoning (\emph{Ilm~al-Maw\={a}r\={\i}th})~\cite{QIAS2025}. This subtask challenges LLMs to navigate the intricate and highly structured field of Islamic inheritance law, which is governed by precise jurisprudential rules. The objective is to develop systems capable of comprehending complex inheritance scenarios, accurately identifying eligible and ineligible heirs, applying fixed-share rules (\emph{far\={a}i\d{d}}), managing residuary shares, and addressing advanced cases such as proportional reduction (\emph{\textsuperscript{\textquoteleft}awl}) and redistribution (\emph{radd}), ultimately performing precise calculations to determine final shares~\cite{mohammedi2012sharia, zouaoui2021islamic}.

The intersection of Natural Language Processing (NLP) and legal reasoning, particularly within specialized domains like Islamic law, has garnered increasing attention. Prior research has explored the application of computational methods to analyze legal texts, extract relevant information, and even automate aspects of legal decision-making. However, the unique complexities of Islamic inheritance law, with its intricate rules and diverse scenarios, present distinct challenges for traditional NLP approaches \cite{Malhas2022QuranQA, Malhas2023QuranQA}.

Recent advancements in Large Language Models (LLMs) have shown promising capabilities in complex reasoning tasks, including those requiring domain-specific knowledge. Studies have demonstrated LLMs' ability to understand and generate human-like text, perform question answering, and even engage in logical inference. However, their performance in highly specialized and rule-based domains often necessitates fine-tuning or integration with external knowledge sources \cite{Almazrouei2023Falcon,Sengupta2023Jais,Alnefaie2023QuranGPT4,Bari2024ALLAM,Mohammed2025AFTINA}.

Specifically, in the context of Islamic inheritance reasoning, several works have emerged \cite{Akkila2016ExpertSystem, Tabassum2019Farayez, zouaoui2021islamic}. For instance, \cite{bouchekif2024inheritance} assesses LLMs on Islamic legal reasoning, providing evidence from inheritance law evaluation. This work highlights the potential and limitations of current LLMs in this domain, underscoring the need for more robust and accurate systems. 

Furthermore, the concept of Retrieval-Augmented Generation (RAG) has gained prominence as a method to enhance LLM performance by grounding their responses in retrieved factual information. This approach is particularly relevant for domains where accuracy and adherence to specific rules are paramount, as it allows LLMs to access and incorporate up-to-date or domain-specific knowledge that may not have been fully captured during their initial training. The integration of RAG with fine-tuned LLMs represents a significant step towards building more reliable and interpretable AI systems for complex reasoning tasks \cite{Alan2024MufassirQAS, Sayeed2025Agentic}.


Our work builds upon these foundations by specifically addressing the challenges of Islamic inheritance reasoning within the framework of a shared task. By combining parameter-efficient fine-tuning with a Retrieval-Augmented Generation (RAG) pipeline, we aim to demonstrate a robust and effective approach for tackling this specialized legal domain, contributing to the broader discourse on applying advanced NLP techniques to complex, rule-governed knowledge systems.

\section{Research Methodology}

Our research methodology for QIAS 2025 SubTask 1 involved a comprehensive approach to address the complexities of Islamic inheritance reasoning using Large Language Models. This section details the task definition, dataset characteristics, the models employed, and our training and inference setup.

\subsection{Task: Islamic Inheritance Reasoning (\emph{Ilm al-Mawārīth})}

SubTask 1 of QIAS 2025 focuses on evaluating the capabilities of LLMs in understanding and reasoning within Islamic inheritance law. The subTask is framed as a multiple-choice question (MCQ) classification problem, where each question has exactly one correct answer. Questions are categorized into two difficulty levels with balanced representation: Beginner (identifying eligible heirs, basic shares, and non-eligible heirs) and Advanced (dealing with multiple heirs, addressing multi-generational cases, fixed estate constraints, and intricate fractional distributions). 

The dataset provided for SubTask 1 consists of a total of 22,000 examples, split into 20,000 examples for model training and 1,000 examples for each validation and testing datasets. Each example is an MCQ related to Islamic inheritance, with question text and up to six answer options (A–F). 


\subsection{Models}

We finetune our primary model \textbf{Fanar-1-9B-Islamic-Inheritance-Reasoning}\footnote{available on HuggingFace:\url{https://huggingface.co/msmadi/Fanar-1-9B-Islamic-Inheritance-Reasoning}} based on \textbf{Fanar-1-9B}\footnote{available on HuggingFace:\url{https://huggingface.co/QCRI/Fanar-1-9B}}, a 9-billion parameter causal decoder-only transformer specifically designed for Arabic and Islamic domain text \cite{fanarllm2025}. 

In addition to the fine-tuned Fanar-1-9B, we integrated it into a \textbf{Retrieval-Augmented Generation (RAG)} pipeline \cite{lewis2020rag} for inference. The RAG setup utilizes the \texttt{all-MiniLM-L6-v2}\footnote{available on HuggingFace:\url{https://huggingface.co/sentence-transformers/all-MiniLM-L6-v2}} embedding model as a retriever to encode questions and retrieve top-$k$ relevant passages  from a \textbf{FAISS} index \cite{8733051, douze2024faiss}. These retrieved passages are then combined with the question and options to form an enriched Arabic chat prompt, which is fed to the fine-tuned Fanar-1-9B model.

\subsection{Training Setup}

Our training setup focused on parameter efficiency and memory optimization. To adapt Fanar-1-9B  LLM efficiently for our task, we employed \textbf{Low-Rank Adaptation (LoRA)} \cite{hu2022lora}. LoRA injects trainable rank-decomposition matrices into specific layers while keeping the original weights frozen. This significantly reduces the number of trainable parameters and computational cost. We also applied \textbf{4-bit NormalFloat (NF4) quantization} \cite{dettmers2022llm} to reduce GPU memory consumption and enabled \textbf{gradient checkpointing} \cite{pytorch2023checkpoint} to reduce peak memory usage. The attention implementation was set to \emph{eager} for improved training stability, and \texttt{use\_cache} was disabled when gradient checkpointing was enabled. Table \ref{tab:finetune-hparams}, provides the key hyperparameters used during model fine-tuning.

\begin{table}[h]
\centering
\begin{tabular}{ll}
\hline
\textbf{Hyperparameter} & \textbf{Value} \\ \hline
Epochs & 4 \\ 
Batch size (per device) & 2 (train and eval) \\ 
Gradient accumulation steps & 32 \\ 
Learning rate & $3\times10^{-4}$ \\ 
Weight decay & 0.01 \\ 
Warmup ratio & 0.1 \\ 
Max gradient norm & 1.0 \\ 
Optimizer & \texttt{adamw\_torch} \\ 
Scheduler & Cosine decay \\
Precision & FP16 \\ \hline
\end{tabular}
\caption{Key hyperparameters for fine-tuning.}
\label{tab:finetune-hparams}
\end{table}

Training data were serialized as \emph{system–user–assistant} turns, where the assistant’s target output is a single gold letter (A–F). LoRA adapters are applied to attention projection and MLP modules (\texttt{q\_proj}, \texttt{k\_proj}, \texttt{v\_proj}, \texttt{o\_proj}, \texttt{gate\_proj}, \texttt{up\_proj}, \texttt{down\_proj}) with $r=32$, $\alpha=64$, and dropout of $0.1$.

For the RAG pipeline, the retrieval $k$ was set to 5, meaning the top 5 relevant passages were retrieved. The maximum input length for the RAG inference was 10,000 tokens, and the maximum new tokens generated by the model was 15. A low temperature of 0.05 was used for decoding, along with a greedy decoding strategy to ensure short, deterministic outputs. Answer extraction was performed using a regex-based procedure to select a single choice letter (A–F).

\section{Evaluation and Results}
For the evaluation of our methodology, we compare our final test results with results reported by the task organizers in \cite{bouchekif2024inheritance, QIAS2025} for testing LLMs with zero-shot prompting on the same test set. The evaluation metric for this task is accuracy. 

\begin{table}[h]
\centering
\begin{tabular}{p{2cm}lccc}
\hline
\textbf{Model} & \textbf{Overall} & \textbf{Beginner} & \textbf{Advanced} \\
\hline
o3         & 93.4 & 94.4 & 92.4 \\
Gemini 2.5 & 90.6 & 91.6 & 89.6 \\
\textbf{QU-NLP} & \textbf{85.8} & \textbf{74.0} & \textbf{97.6} \\
GPT-4.5    & 74.0 & 86.8 & 61.2 \\
LLaMA3     & 48.8 & 57.8 & 39.8 \\
Fanar 7B   & 48.1 & 60.4 & 35.8 \\
Mistral    & 44.5 & 58.6 & 30.4 \\
ALLaM7B    & 42.9 & 58.0 & 27.8 \\
\hline
\end{tabular}
\caption{Accuracy (\%) for each model across difficulty levels. Other models results are based on zero-shot setting using Arabic prompts as reported in \cite{bouchekif2024inheritance, QIAS2025} }
\label{tab:model-accuracy}
\end{table}

As presented in Table \ref{tab:model-accuracy}, QU-NLP, achieved an overall accuracy of 85.8\%, outperforming other competitive models such as, GPT 4.5,  LLaMA 3 70B\footnote{Available via the Groq API: \url{https://console.groq.com/keys}}, Fanar (Islamic-RAG\footnote{Available via a free public API: \url{https://api.fanar.qa/request/en}}),  Mistral-Saba-24B\footnote{Available via the Groq API: \url{https://console.groq.com/keys}} and ALLaM-7B\footnote{Arabic model hosted on Hugging Face: \url{https://huggingface.co/Abdelaali-models/ALLaM-7B-Instruct-preview}} and achieving competitive results behind state of the art commercial LLMs in reasoning capabilities, such as: Gemini 2.5 (flash-preview), OpenAI’s o3. While our system did not achieve the top rank, QU-NLP (with RAG) surpassed all models on the advanced subset of the testing dataset (500 MCQs) with accuracy of 97.6\%.  This result demonstrates the effectiveness of our approach, which combines LoRA fine-tuning of the Fanar-1-9B model with a Retrieval-Augmented Generation (RAG) pipeline, in addressing the complex reasoning challenges posed by Islamic inheritance law. Our model's performance indicates a strong capability in comprehending inheritance scenarios, identifying heirs, and applying the intricate rules required for accurate share calculation.

\section{Discussion}

We evaluate a multiple-choice inheritance reasoning system on 1{,}000 items with an overall accuracy of 85.8\%. Performance differs sharply by level: \emph{Beginner} = 74.0\% (n=500) vs.\ \emph{Advanced} = 97.6\% (n=500). Two phenomena account for most residual errors at the Beginner level. First, items whose correct answer indicates a \textAR{محجوب} (``blocked'') heir are substantially harder (64.5\%, n = 299) than all other cases (94.9\%, n = 701), suggesting the model sometimes assigns shares despite the presence of higher-priority heirs. Second, questions containing explicit negation or exception cues (e.g., \textAR{لا}/\textAR{ليس}/\textAR{لم}/\textAR{لن}/\textAR{غير}/\textAR{بدون}) yield lower accuracy (83.5\%, n = 807) compared to those without negation (95.3\%, n = 193), indicating occasional polarity flips.

To further investigate QU-NLP’s limitation on blocked cases, we analyzed the count of questions whose gold answer is \textAR{محجوب} in the development and training splits. We found that blocked items constitute only 1.70\% of development set (17/1{,}000) but 17.46\% of train (3{,}491/20{,}000), whereas (for reference) they account for 29.90\% of Test (299/1{,}000). This mismatch—especially the severe under-representation in Development set—helps explain the degraded Test performance on blocked questions (64.55\% vs.\ 94.86\% on non-blocked).

A further class of errors results from near-duplicate answer options where orthographic differences (e.g., \textAR{باقى} vs.\ \textAR{باقي}) leave the semantics unchanged but map to different label IDs. We found 10 such cases (about 7\% of all errors). These are dataset artifacts rather than modeling deficiencies. After normalizing Arabic orthography (removing diacritics and unifying letter forms), gold and predicted options collapse to the same string. For transparency, Appendix ~\ref{sec:appendix} lists two misclassified examples across the three categories: (A) blocked heirs (\textAR{محجوب}), (B) negation/exception cues, and (C) near-duplicate option texts, and Table ~\ref{tab:error-counts} demonstrates the counts of misclassified questions per category of error and level.

\begin{table}[h]
\centering

\begin{tabular}{p{2cm}lrrr}
\hline
\textbf{Category} & Advanced & Beginner & Total \\
\hline
Blocked (\textAR{محجوب}) & 0 & 106 & 106 \\
Negation-Exception & 3 & 14 & 17 \\
Near-duplicate options & 0 & 10 & 10 \\
Other & 9 & 0 & 9 \\
\hline
\textbf{All errors} & 12 & 130 & 142 \\
\hline
\end{tabular}
\caption{Misclassification counts by category and level (total errors = 142).}
\label{tab:error-counts}
\end{table}

To mitigate these errors, we suggest: (i) adding explicit post-rules or contrastive training focused on hijb (\textAR{محجوب}) cases; (ii) augmenting training with negation/exception rewrites; and (iii) normalizing and deduplicating answer options during dataset curation and evaluation to avoid orthography-induced label mismatches.

\section{Conclusion}

This paper presented our system, QU-NLP, for SubTask 1: Islamic Inheritance Reasoning at the QIAS 2025 Shared Task. We demonstrated the application of a LoRA fine-tuned Fanar-1-9B causal language model integrated within a Retrieval-Augmented Generation (RAG) pipeline to address the intricate challenges of Islamic inheritance law. Our methodology focused on parameter-efficient fine-tuning and leveraging external knowledge retrieval to enhance the model's reasoning capabilities and factual accuracy in this specialized domain.

Our system achieved an accuracy of 0.858 in the final test, securing a competitive position among the participants. This result highlights the significant potential of combining advanced LLM architectures with retrieval mechanisms for complex, rule-based legal reasoning tasks. We successfully navigated challenges related to memory constraints through techniques like 4-bit NF4 quantization and gradient checkpointing, making the deployment of such large models more feasible.

Future work will explore further enhancements to the RAG pipeline, including more sophisticated retrieval strategies and the potential incorporation of explicit symbolic reasoning components to handle the highly structured nature of Islamic jurisprudence. Additionally, investigating methods for generating interpretable justifications for the model's predictions could provide deeper insights into its reasoning process and build greater trust in its applications.

\bibliography{custom}

\appendix

\section{Misclassified Examples}
\label{sec:appendix}

\begin{table*}[h]
\centering
\footnotesize

\label{tab:combined-misclass}
\begin{tabular}{p{2.1cm} p{2cm} p{6.1cm} p{6.2cm}}
\hline
\textbf{Category} & \textbf{ID} & \textbf{Question (excerpt)} & \textbf{Gold / Predicted} \\
\hline
Blocked (\textAR{محجوب}) & 9337\_nf5j2z5o\_6 &
\textAR{مات وترك: بنت ابن ابن (2) و بنت (4) و ابن ابن عم لأب و زوجـة و أخ لأب (3) و أخ شقيق (3) كم النصيب الأصلي لـ ابن ابن عم لأب من التركة، وما الدليل على ذلك؟} &
(C)\, \textAR{نصيبه هو محجوب، والدليل: لا يرث ابن ابن عم لأب فى وجود الفرع الوارث المذكر - مثل الإبن أو ابن الإبن وإن نزل - ولا الأصل المذكر - مثل الأب وأب الأب وإن علا- ولا فى وجود الإخوة الأشقاء أو لأب ولا عند إجتماع الأخت مع أحد البنات} \\
&&&
(D)\, \textAR{نصيبه هو لا شيء، والدليل: لا يرث ابن ابن عم لأب فى وجود الفرع الوارث المذكر - مثل الإبن أو ابن الإبن وإن نزل - ولا الأصل المذكر - مثل الأب وأب الأب وإن علا- ولا فى وجود الإخوة الأشقاء أو لأب ولا عند إجتماع الأخت مع أحد البنات} \\[2pt]

Blocked (\textAR{محجوب}) & 1245\_nn7z0t6w\_1 &
\textAR{مات وترك: ابن ابن أخ لأب (4) و أخ شقيق (2) و عم الأب لأب (2) و ابن عم شقيق (4) و أم الأب كم النصيب الأصلي لكل صنف من الورثة من التركة؟} &
(F)\, \textAR{أم الأب: السدس، أخ شقيق (2): باقى التركة، ابن ابن أخ لأب(4): محجوب، عم الأب لأب(2): محجوب، ابن عم شقيق(4): محجوب} \\
&&&
(A)\, \textAR{أم الأب: السدس، أخ شقيق (2): باقى التركة، ابن ابن أخ لأب(4): عصبة، عم الأب لأب(2): محجوب، ابن عم شقيق(4): محجوب} \\[4pt]

Negation/Exception & 3818\_ne5o6t0g\_2 &
\textAR{مات وترك: أخت شقيقة (3) و أخت لأم (2) و ابن أخ لأب (2) كم النصيب الأصلي لـ أخت شقيقة (3) من التركة، وما الدليل على ذلك؟} &
(F)\, \textAR{نصيبه هو الثلثان، والدليل: الأخت الشقيقة - عند عدم الأخ الشقيق - مثلها مثل البنت - إذا لم يكن هناك بنات صلبيات أو بنات ابن - فتأخذ الشقيقة النصف ان كانت واحده والثلثان ان كانتا اثنتين أو أكثر ... وإلا حجبت بهم} \\
&&&
(B)\, \textAR{نصيبه هو كل التركة، والدليل: الأخت الشقيقة - عند عدم الأخ الشقيق - مثلها مثل البنت ... وإلا حجبت بهم} \\[4pt]

Negation/Exception (in explanation) & 877\_nr5a8q3s\_2 & \textAR{مات وترك: بنت ابن (3) و أخ لأب (2) و ابن أخ لأب (4) و أب الأب و ابن عم لأب (2) و ابن عم الأب (3) كم النصيب الأصلي لـ بنت ابن (3) من التركة، وما الدليل على ذلك؟} & (E)\, \textAR{نصيبه هو الثلثان، والدليل: بنات الإبن - مثل بنت الإبن وبنت ابن الإبن - مثلهن مثل البنت بشرط عدم وجود بنت صلبيه أوابن صلبى أو ابن ابن أعلى منهن فيحجبهن. فترث الواحدة من بنات الابن النصف إذا لم يكن هناك ابن ابن فى درجتها يعصبها وترث الأكثر من واحدة الثلثين . قال تعالى (يُوصِيكُمُ اللَّهُ فِي أَوْلادِكُمْ لِلذَّكَرِ مِثْلُ حَظِّ الأُنثَيَيْنِ فَإِنْ كُنَّ نِسَاءً فَوْقَ اثْنَتَيْنِ فَلَهُنَّ ثُلُثَا مَا تَرَكَ وَإِنْ كَانَتْ وَاحِدَةً فَلَهَا النِّصْفُ)} \\
&&& (A)\, \textAR{نصيبه هو لا شيء، والدليل: بنات الإبن - مثل بنت الإبن وبنت ابن الإبن - مثلهن مثل البنت بشرط عدم وجود بنت صلبيه أوابن صلبى أو ابن ابن أعلى منهن فيحجبهن. فترث الواحدة من بنات الابن النصف إذا لم يكن هناك ابن ابن فى درجتها يعصبها وترث الأكثر من واحدة الثلثين . قال تعالى (يُوصِيكُمُ اللَّهُ فِي أَوْلادِكُمْ لِلذَّكَرِ مِثْلُ حَظِّ الأُنثَيَيْنِ فَإِنْ كُنَّ نِسَاءً فَوْقَ اثْنَتَيْنِ فَلَهُنَّ ثُلُثَا مَا تَرَكَ وَإِنْ كَانَتْ وَاحِدَةً فَلَهَا النِّصْفُ)} \\[6pt]

Near-duplicate options & 8804\_nl1d9s7s\_4 &
\textAR{مات وترك: عم الأب لأب (4) و أخت لأب (5) و عم لأب (2) و أم أم الأب و أم أم الأم كم النصيب الأصلي لـ عم لأب (2) من التركة، وما الدليل على ذلك؟} &
(B)\, \textAR{نصيبه هو \textbf{باقى} التركة، والدليل: لأنه عصبة} \\
&&&
(E)\, \textAR{نصيبه هو \textbf{باقي} التركة، والدليل: لأنه عصبة} \\[2pt]

Near-duplicate options & 4434\_nr1f0y8b\_4 &
\textAR{مات وترك: أب أب الأب و أخت لأب (5) و عم الأب (5) و أم الأب كم النصيب الأصلي لـ أخت لأب (5) من التركة، وما الدليل على ذلك؟} &
(A)\, \textAR{نصيبه هو \textbf{باقى} التركة، والدليل: لأنه عصبة} \\
&&&
(C)\, \textAR{نصيبه هو \textbf{باقي} التركة، والدليل: لأنه عصبة} \\
\hline
\end{tabular}
\caption{Illustrative misclassified examples across three categories: (A) blocked heirs (\textAR{محجوب}), (B) negation/exception cues, and (C) near-duplicate option texts.}
\end{table*}

\end{document}